\documentclass[conference]{IEEEtran}
\IEEEoverridecommandlockouts
\usepackage{cite}
\usepackage{balance}
\usepackage{amsmath,amssymb,amsfonts}
\usepackage{algorithmic}
\usepackage{graphicx}
\usepackage{textcomp}
\usepackage{xcolor}
\def\BibTeX{{\rm B\kern-.05em{\sc i\kern-.025em b}\kern-.08em
    T\kern-.1667em\lower.7ex\hbox{E}\kern-.125emX}}
\begin{document}

\title{Deep Neural Networks on EEG Signals to Predict Auditory Attention Score Using Gramian Angular Difference Field \\
}

\makeatletter
\newcommand{\linebreakand}{%
  \end{@IEEEauthorhalign}
  \hfill\mbox{}\par
  \mbox{}\hfill\begin{@IEEEauthorhalign}
}
\makeatother

\author{

\IEEEauthorblockN{ Mahak Kothari\textsuperscript{1}, Shreyansh Joshi\textsuperscript{1}, Adarsh Nandanwar\textsuperscript{1}, Aadetya Jaiswal\textsuperscript{1}, Veeky Baths\textsuperscript{2}}
\IEEEauthorblockA{
\textit{\textsuperscript{1}Department of Computer Science \& Information Systems}\\
\textit{\textsuperscript{2}Cognitive Neuroscience Lab, Department of Biological Sciences} \\
\textit{BITS Pilani K K Birla Goa Campus, Goa, India}\\
\{f20180232, f20180097, f20180396, f20180096, veeky\}@goa.bits-pilani.ac.in
}

% \IEEEauthorblockN{ Mahak Kothari}
% \IEEEauthorblockA{\textit{Dept. of Computer Science \&}\\
% \textit{Information Systems} \\
% \textit{BITS Pilani K K Birla Goa Campus}\\
% Goa, India\\
% f20180232@goa.bits-pilani.ac.in}
% \and
% \IEEEauthorblockN{Shreyansh Joshi}
% \IEEEauthorblockA{\textit{Dept. of Computer Science \&}\\
% \textit{Information Systems} \\
% \textit{BITS Pilani K K Birla Goa Campus}\\
% Goa, India\\
% f20180097@goa.bits-pilani.ac.in}
% \and
% \IEEEauthorblockN{Adarsh Nandanwar}
% \IEEEauthorblockA{\textit{Dept. of Computer Science \&}\\
% \textit{Information Systems} \\
% \textit{BITS Pilani K K Birla Goa Campus}\\
% Goa, India\\
% f20180396@goa.bits-pilani.ac.in}
% \linebreakand
% \IEEEauthorblockN{Aadetya Jaiswal}
% \IEEEauthorblockA{\textit{Dept. of Computer Science \&}\\
% \textit{Information Systems} \\
% \textit{BITS Pilani K K Birla Goa Campus}\\
% Goa, India\\
% f20180096@goa.bits-pilani.ac.in}
% \and
% \IEEEauthorblockN{Veeky Baths}
% \IEEEauthorblockA{\textit{Cognitive Neuroscience Lab}\\
% \textit{Department of Biological Sciences} \\
% \textit{BITS Pilani K K Birla Goa Campus}\\
% Goa, India\\
% veeky@goa.bits-pilani.ac.in}

}

\maketitle

\begin{abstract}
Auditory attention is a selective type of hearing in which people focus their attention intentionally on a specific source of a sound or spoken words whilst ignoring or inhibiting other auditory stimuli. In some sense, the auditory attention score of an individual shows the focus the person can have in auditory tasks. The recent advancements in deep learning and in the non-invasive technologies recording neural activity beg the question, can deep learning along with technologies such as electroencephalography (EEG) be used to predict the auditory attention score of an individual? In this paper, we focus on this very problem of estimating a person's auditory attention level based on their brain's electrical activity captured using 14-channeled EEG signals. More specifically, we deal with attention estimation as a regression problem. The work has been performed on the publicly available Phyaat dataset. The concept of Gramian Angular Difference Field (GADF) has been used to convert time-series EEG data into an image having 14 channels, enabling us to train various deep learning models such as 2D CNN, 3D CNN, and convolutional autoencoders. Their performances have been compared amongst themselves as well as with the work done previously. Amongst the different models we tried, 2D CNN gave the best performance. It outperformed the existing methods by a decent margin of 0.22 mean absolute error (MAE).\\
\end{abstract}


\section{Introduction}

Attention plays a major role in our daily lives. Almost every conscious move we make requires attention. Clearly, understanding attention and improving it can be the key to ameliorate the effectiveness of many tasks. Auditory attention is a form of attention, a cognitive process wherein the attention of the person is gauged by listening/hearing tasks. More specifically, it refers to the process by which a person focuses selectively on a stimulus of interest whilst ignoring other stimuli. It is important to note that such selective hearing is not some kind of a physiological disorder like autism, dementia but rather, it is the ability of humans to intentionally block out certain sounds and noise and to focus on selective ones. 

The next obvious question that arises is that why is auditory attention relevant in the real world? Sounds in everyday life seldom appear in isolation. Both humans and machines are constantly flooded with a cacophony of sounds that need to be filtered and scoured for relevant information— famously referred to as the 'cocktail party problem'.  

This paper makes a conscious effort towards predicting the auditory attention level of an individual by analyzing the EEG signals produced by the person's brain. Since the world is moving towards automation in almost every domain these days, taking a step in that direction by automating the prediction of the attention score of a person, just by using EEG signals from the brain, was a big source of motivation for our work. 

We, in this paper, have tried to show that level of attention (attention score) can be predicted based on the EEG signals recorded from the brain. EEG is a non-invasive technique used to measure the electrical activity of the brain regions~\cite{KUMAR2012, Vaid2015}. A lot of research on attention has been done in the past using EEG signals. Studies have shown that EEG data from different frequency bands can be used to draw inferences on visual and auditory attention. In 2006, Sauseng et al.~\cite{Sauseng2006} showed that shift of visual-spatial attention is selectively associated with human EEG alpha activity. In 2013, Gola et al.~\cite{GOLA2013} showed that EEG beta-band activity is related to attention and attentional deficits in the visual performance of elderly subjects. Decreased activity of beta-band reflected the difficulty in activation of attentional processes (alertness deficits in short delay condition) and deficits in sustaining those processes (longest delay). As recently as in 2018, Wang et al.~\cite{wang2018} used deep learning techniques like convolutional neural networks to work on visual attention. This method was based on an end-to-end deep learning architecture, which resulted in good performance while predicting human eye fixation with view-free scenes.

However, unlike visual attention, auditory attention analysis based on EEG signals is a relatively lesser-explored and newer field, although some efforts have been made to analyze auditory attention in the past using EEG signals. In 2014 Treder et al.~\cite{Treder2014} showed that by observing the EEG signals, we could infer the attention paid to a particular instrument in polyphonic music. With a mean accuracy of 91\%, they classified the attended instrument. Another study in 2017 by Fiedler et al.~\cite{Fiedler2017} showed that using the EEG response recorded from short-distance configurations consisting only of a single in-ear-EEG electrode and an adjacent scalp-EEG electrode, an individual's attention focus could be detected. This information can be used in hearing aids to identify the listener's focus of attention in concurrent-listening ('cocktail party') scenarios. Similarly, by analyzing the electrode potential, we may be able to estimate the auditory attention level.
How noise affects the attention levels of an individual was shown by Das et al.~\cite{Das_2018}. They showed that it is possible to identify the speaker to which the listener is actively listening in an environment with multiple speakers. They observed that attention in moderate background noise is greater than in a no-background noise environment. A further increase in the noise was found to bring about a remarkable drop in attention level. In our work, the EEG data source~\cite{bajaj2020} features recordings at 6 different signal-to-noise ratios (SNR). This noise data is crucial in accurately estimating auditory attention levels.

In this paper, we aim to estimate auditory attention using deep learning architectures like 2D CNN, 3D CNN, and convolutional autoencoders with random forest and XGBoost regressor. This is not the first work in the field involving convolutional neural networks (CNN)~\cite{lecun1998}. In~\cite{Deckers2018}, various CNN-based approaches were used to decode attention and detect the attended speaker in a multi-speaker scenario. The majority of time-series classification work on signals till date (involving CNNs) has been done using 1D CNN. Kashiparekh et al.~\cite{Kashi2019}, Fawaz et al.~\cite{Fawaz2019} and Wang et al.~\cite{Wang2017} succeeded in demonstrating the significant results of 1D CNN. However, a major limitation in using 1D CNN is the extensive hyperparameter tuning accompanying it~\cite{WTang2020}. Therefore, we restrained ourselves from using 1D CNN. Recently, 2D CNNs have been applied to time-series data by transforming them into 3D images. Hatami et al.~\cite{Debayle2018} were among the very first researchers to propose encoding of the time-series data into images using Gramian Angular Fields (GAF) and then applied the CNN-based architecture. GAF alongside CNN~\cite{Bragin2019}, has also been previously used for the classification of motion imagery from EEG signals with high accuracy. In~\cite{Krishnan2020}, the EEG time-series data were transformed to RGB images using Gramian Angular Summation Field (GASF), and various deep neural networks were used to diagnose epilepsy. Using a similar method, we plan to transform the time-series data into multi-channel images using Gramian Angular Difference Field (GADF).\\
The remaining part of the paper is organized as follows. Section 2 and 3 talk about the dataset on which the work has been done and data preprocessing using GADF, respectively. Section 4 contains the methodology, which is further divided into three subsections. Section 5 talks about the evaluation metric used. Section 6 presents the results and inferences we obtained from the various experiments we performed, and finally, section 7 wraps up the paper with the conclusion and future work.

\section{Dataset}

In this paper, we have used the PhyAAt dataset~\cite{bajaj2020}, which contains the physiological responses of 25 subjects collected from an experiment on auditory attention. The chosen subjects were university students aged between 16 to 34. All participants were non-native English speakers, out of which 21 of them were male, and 4 were female. 

The following procedure was performed to obtain the dataset. Each participant was subjected to 3 tasks - listening, writing, and resting, across multiple trials. More formally, the experiment consisted of 144 trials for each subject, where each trial was divided into the 3 aforementioned tasks. During each listening phase, an audio clip was played that contained English language sentences of varying lengths (ranging from 3 to 13 words) with 6 different levels of background noise. A good chunk of the sentences were semantic in nature, while others were non-semantic. 

During each writing phase, the subject was asked to write the words in the correct order. The attention score was calculated using the number of words present in the same positions in the listening and writing phase for a given sentence. The formula for the same has been given in Eq.~\ref{attention}
\begin{equation}
\label{attention}
    Attention\:score = \frac{number\:of\:correct\: words}{total \:number\: of\: words}
\end{equation}

A 14-channel EEG system was used to record the brain signals of the subjects during the experiment at a sampling rate of 128Hz. A total of 144 stimuli were collected for each participant. In our work, we have used signals only from the listening phase for the prediction.

Phyaat dataset is one of the best publicly available datasets for determining auditory attention using EEG signals. Albeit small in size, the diversity of subjects, as well as the rigour with which the experiments were performed (for instance, to better estimate the auditory attention, the experiment also included some non-semantic English sentences as well as 6 different levels of background noise) increases the possibility of a model to generalize well on real-world data, after being trained on it. 

\section{Data pre-processing using GAF}

The deep learning boom is largely fueled by the success that deep learning based models have had in computer vision~\cite{Voulodimos2018} and speech recognition~\cite{Nassif2019, Hannun2014}. The well-established success of CNNs in feature extraction from images was one of our main motivations for choosing computer vision based models over RNNs and LSTMs that are typically used while dealing with time-series data. To do that, we had to first convert the time-series data into images, for which we have used the technique of Gramian Angular Field (GAF).

The dataset contained the time-series data of EEG signals, which was converted into a GAF image~\cite{wang2015_encoding}. GAF is simply a polar coordinate representation of the time-series data, which preserves the temporal dependency. Gramian Difference Angular Field (GADF) is a type of GAF in which trigonometric difference between each point is calculated to create a matrix with the temporal correlation amongst different time intervals.

Given below is the step-by-step procedure, we followed to create a GADF image from the EEG signals obtained from a particular electrode. Fig.~\ref{timeseries}. visually depicts that procedure. \\
\begin{itemize}
\item Step 1: Let $X$ (Eq.~\ref{X}) be a time-series datum containing $k$ real values. Apply a rescaling function $f$ (Eq.~\ref{func}) with domain ${\Bbb R}$ and range [-1,1]. This is essential so as to apply $arccos$ function, while converting them into polar coordinates as shown in step 2.\\
\begin{equation}
\label{X}
    X = \{x_1, x_2, ... , x_k\}\\
\end{equation}
\begin{equation}
\label{func}
    f(x_i) = \frac{(x_i - max(X) ) + ( x_i-min(X) )}{max(X) - min(X) }
\end{equation}
\\

\item Step 2: Convert the rescaled data into polar coordinates by encoding the value as the angular cosine $\phi$ and the time stamp as the radius $r$, as can be seen below. \\
% Here $r$ is the radius and $\phi$ is the angle in the polar coordinate system.\\
\\
\(
\begin{cases}
  \phi_i = arccos(f(x_i)),\: 1 \leq i \leq k \\
  r = \frac{i}{k}
\end{cases}
\)
\\
\\
\item Step 3: Calculate the GADF matrix\\

    \(GADF =
    \begin{bmatrix}
    \sin(\phi_1 - \phi_1) & ..  & \sin(\phi_1 - \phi_k)\\
    \sin(\phi_2 - \phi_1) & ..  & \sin(\phi_2 - \phi_k)\\
    : & : & :\\
    \sin(\phi_k - \phi_1)& ..  & \sin(\phi_k - \phi_k)\\
    \end{bmatrix}
    \)
    \\\\
\end{itemize}
\begin{figure}
\includegraphics[width=9.5cm]{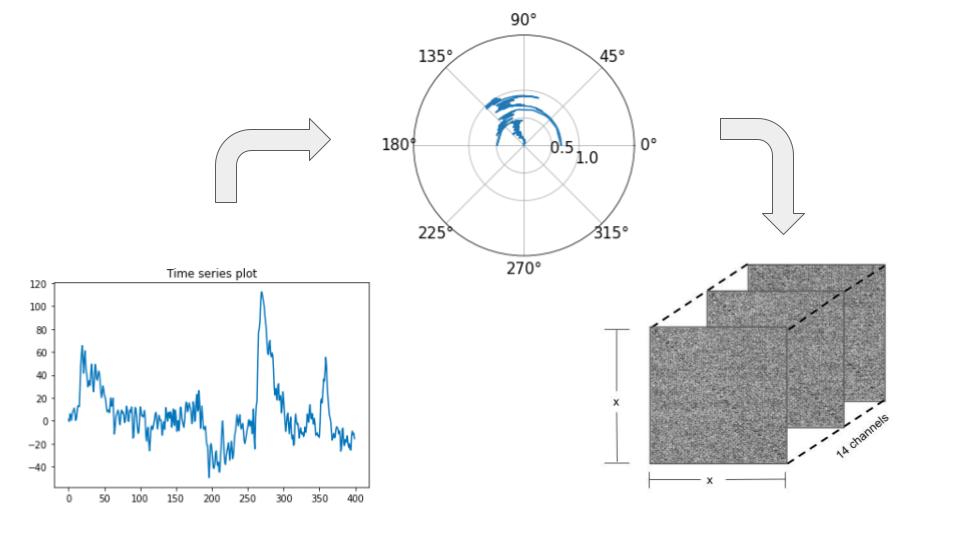}
\caption{Time-series to GADF conversion.}
\label{timeseries}
\end{figure}

The above-mentioned procedure was followed for the EEG signals from all the 14 electrodes planted in a subject's head, resulting in 14 images which were eventually stacked one behind the other, resulting in a single 14-channel GADF image. However, these images were not of fixed dimensions. A way to deal with this was to pad all those images so that they become of the same size. However, we refrained from doing so as the padding might result in the loss of information from some images, possibly due to the variability in image sizes. Instead, the models were designed in a way that they could incorporate images of different sizes.

\section{Methodology}

For a total of 25 subjects, we had about 3600 datapoints (144x25, 144 trials each for 25 subjects) available for our work, out of which 300 datapoints were chosen to be kept for validation, and the remaining 90\% of the data would be used for training purposes. It is worthwhile to note that this training and testing data was not fixed during the entire duration of our experiment. The dataset (consisting of 3600 data points as mentioned previously) was randomly divided into 12 sets or 'folds', and experiments were conducted by choosing one fold at a time as the validation set, while the rest of the 11 folds served as the training set. 

This technique was employed to ensure more reliable and reproducible results since not much work has been done in this field (on the Phyaat dataset) to compare our work with. The value of 12 (for the number of folds) was chosen after careful deliberation and taking into account the variance and standard deviation of the results. A higher number of folds imply a larger computation time and a greater variance, as each validation fold has lesser datapoints. On the other hand, an extremely smaller number of folds is detrimental in the way that a larger chunk of data would not be available for training, thus leading to underfitting, with the model not being able to properly discern the patterns present in the dataset. The sweet spot for the number of folds was found by experimenting with different values. It was found that any value between 11 to 13 was giving similar results. We chose 12, since it divides 3600 completely and so we could ensure that all datapoints get a chance to be a part of validation set. For each of the following architectures, we employed the 12-fold cross-validation and calculated the mean and standard deviation of the MAE.

% As mentioned previously, we have used computer vision based models to solve this regression problem of predicting the attention score. The reason can be ascribed to the well-established feature extracting ability of CNNs from images. In recent times, the uses of CNNs have diversified even more, and they are not just limited to images.

We trained and tested the following computer vision based models - 

\subsection{2D CNN}

A 2D CNN is capable of only extracting spatial features in images whilst neglecting the temporal correlation since the kernel moves only in the spatial directions. The architecture we used comprises 4 blocks of convolution layers followed by maxpooling to reduce the dimensions as the depth of the network increases. The number of filters increase with the depth, but the dimensions of filters in both Conv2d layers, as well as the MaxPool2d layers, remain the same ((3,3) and (2,2) respectively). The activation function ReLU was used between the layers, as studies have proven empirically that deep networks, when training with ReLU, tend to converge quickly and more reliably as against any other activation function.

Generally, a CNN works on the image of a fixed dimension because the fully connected layer requires a specific number of neurons, but in our case, as explained previously, the dimensions of input images were not fixed, so in the last of the 4 blocks described above, we introduced an adaptive maxpooling layer (instead of the normal maxpooling) before flattening the representation. 'AdaptiveMaxPool2d' converts the input of any dimension into the output of a fixed dimension. Following it, our architecture has 3 dense layers with 128, 16 and 1 neurons. Dropout was used in the first of the 3 FC layers to combat overfitting. We chose against using dropout in the convolutional layers as those are the layers responsible for learning the features, and hence we did not want any loss of useful information during the feature learning process. Fig.~\ref{arch} contains the detailed architecture. 
% \begin{figure}
% \includegraphics[width=8cm]{arch.png}
% \caption{The architecture of 2D CNN that gave the best result.}
% \label{arch}
% \end{figure}
\\
\begin{figure*}[ht]
\centering
\includegraphics[width=\textwidth]{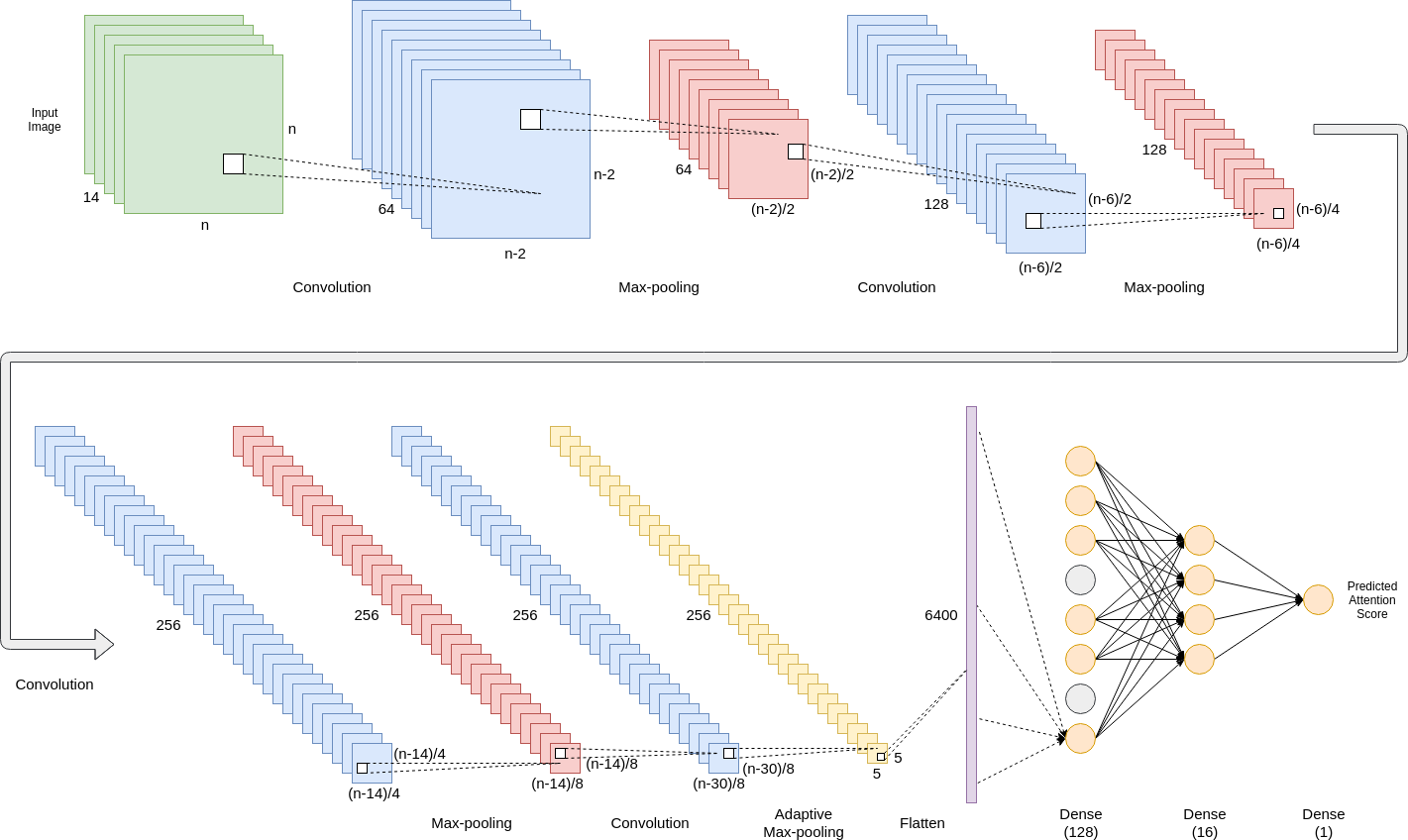}
\caption{The architecture of 2D CNN that gave the best result.}
\label{arch}
\end{figure*}

\subsection{3D CNN}

Generally, 3D CNNs are used to learn the representations for volumetric data like videos where temporal relations among different frames are captured along with spatial relations of each frame. The third axis intrinsically links the frames together and so cannot be ignored. Conventionally, a 3D CNN operates on 4 dimensions of an image - width, height, depth, channels as against 3 in the case of 2D CNN. In contrast to 2D CNNs, where the filters are used only in spatial directions, the 3D CNN uses the filters in a temporal direction also.

Since we were dealing with 14-channel images, we felt this architecture would be ideal for our work. We believed that there might be some relationship amongst the different channels of the EEG signals that can be captured using a 3D CNN (and probably not by 2D CNN). 
Here, we chose to go with a shallower network, as compared to the 2D CNN one, for the simple reason that 3D CNN is very computationally demanding, and we had limited memory resources at hand. This time we had 3 blocks of convolution layers, followed by maxpooling. As in the case of 2D CNN, the number of filters increase with the depth (64,128,128), but the dimensions of filters in both Conv3d layers, as well as the MaxPool3d layers remain the same ((3,5,5) and (1,3,3) respectively). The activation function ReLU was used between the layers. The 3D version of adaptive maxpooling, AdaptiveMaxPool3d was used before flattening the extracted features. Following it, we used 3 dense layers having 512, 16 and 1.

\subsection{Convolutional Autoencoder}

Autoencoder~\cite{HinSal06} is a kind of artificial neural network used to encode and compress data. They work by compressing the input first and later reconstructing the output from the compressed representation of the data.
In this paper, we have used a nine-layered convolutional autoencoder to learn the encoding of the data. Convolutional autoencoder is a variant of CNNs that are used as the tools for unsupervised learning of convolution filters. They are used to preserve the spatial information of an image and work differently from general autoencoders that completely ignore the 2D image structure. Since our inputs consisted of images (after applying GADF), it made sense to use CNN (convnets) as encoders and decoders. To build any autoencoder, three things are required: an encoding function, a decoding function, and a distance function between the amount of information loss between the compressed representation of the input data and the decompressed representation (i.e., a "loss" function). 

Our encoder is comprised of 3 blocks of convolution layers and maxpooling. As in the case of 2D CNN and 3D CNN, the number of filters increase with the depth of the network (64,128,128) and the kernel size remains the same throughout for both Conv2d layers and MaxPool2d layers ((3,3) and (2,2), respectively). The last of the 3 blocks had the MaxPool2d replaced with AdaptiveMaxPool2d to deal with the variable size of input images. The encoder finally terminates with an FC layer with 512 neurons. The decoder was designed in the reverse order as the encoder. It begins with an FC layer, followed by 3 blocks of maxunpooling and transpose convolution. In this case, with the increasing depth of the network, the number of filters keeps on decreasing (128,128,64). The kernel size remains the same throughout for both MaxUnpool2d and ConvTranspose2d ((2,2) and (3,3) respectively). The decoder finally creates an image with the same shape as the input image to the encoder. For both the encoder, as well as the decoder, the activation function ReLU was used between the layers. The distance function used was MAE, which actually calculates the pixel-wise difference between the original input image and the new image generated by the decoder.

The encoded data was passed to two of the most promising machine learning regression models: First, the random forest regressor~\cite{Breiman2001}, which is an ensembling technique that uses the output of multiple decision trees and takes the mean of all of them for the final output. 
Second, the XGBoost regressor~\cite{Chen2016}, a state-of-the-art machine learning algorithm that uses the boosting ensemble technique. In boosting, multiple weak learners are trained sequentially such that each tree tries to reduce the error of the previous one~\cite{Freund1999}. The hyperparameters of both the machine learning models were found by fine-tuning using GridSearchCV.

In all the 3 architectures, the models were trained using Adam optimizer for about 15 epochs. It was observed that training any further, caused the models to overfit. The initial learning rate was fixed at about 0.002-0.003 in each case, and learning rate decay was used to help the model converge and generalize better. Mean Squared Error (MSE) was the loss function used during the training process.

\section{Evaluation Metric}

The closeness of the actual value to the predicted value of the attention score was used to evaluate the performance of our models. The metric we have used for the attention score estimation is the mean absolute error or MAE that captures the average magnitude of error in a set of predictions. 
MAE calculates the mean of the absolute error between actual score and predicted score as defined by the Eq.~\ref{mae}.
\begin{equation}
    MAE = \frac{1}{n}\sum_{j=1}^{n}|y_j - \hat{y_j}| 
    \label{mae}
\end{equation}
Here $n$ is the number of testing samples, $y_j$ denotes the actual attention score, and $\hat{y_j}$ is the predicted score of the j-th sample.

\section{Results and inferences}

Table~\ref{mae-table} compares the performance of different models by displaying the mean and standard deviation of the validation MAE each model obtained, after taking into account the 12-fold cross-validation, we had performed during training. While the mean provides a good measure of how accurately the model predicted the attention score on average, the standard deviation shows the amount of variance in the predicted values over the 12 folds of validation data. A higher variance would typically mean that the trained model is prone to be affected by outliers and might as well, have a lesser generalization ability.

2D CNN was found to provide the best results when it comes to the mean MAE predicted over the validation set. We had expected 3D CNN to perform better than 2D CNN, since it has the potential to capture the relations between the 14 channels of the image. The most plausible explanation for it not performing up to our expectations is the shallowness of the network that we had trained our model on. We had limited memory resources at our disposal and coupled with the 14-channel input image, and it was not possible to train a deeper network that could extract features in a better way. This is something we hope to address in our future work.

\begin{table}[hbt!]
\centering
\caption{Comparision of Different Models}
\begin{tabular}{c c} 
\hline
Models: & Validation MAE (mean $\pm$ std)\\
\hline
\\
\textbf{2D CNN} & \textbf{$\boldsymbol{29.43}\pm \boldsymbol{1.81}$} \\
\\
3D CNN & $29.72 \pm 1.76 $\\
\\
Autoencoder with Random Forest & $30.16 \pm 1.65$\\
\\
Autoencoder with XGBoost & $29.99 \pm 1.91$\\
\\
SVM~\cite{bajaj2020} & $29.65 \pm 4.75$ (approx.)\\ 

\end{tabular}
\label{mae-table}
\end{table}

However, despite the shallowness of the network, it can be seen that 3D CNNs have performed the second-best among the networks, and in fact, almost as good as the owners of the Phyaat dataset were able to achieve. Even though it was not able to perform up to its full potential because of the resource constraints, it managed to outperform a couple of other architectures.
Among the convolutional autoencoders, the better results of the XGBoost model can be ascribed to the fact that we found it to generalize better, thereby overfitting lesser as compared to the model with random forest.
\\
\begin{figure}
\includegraphics[width=9.5cm]{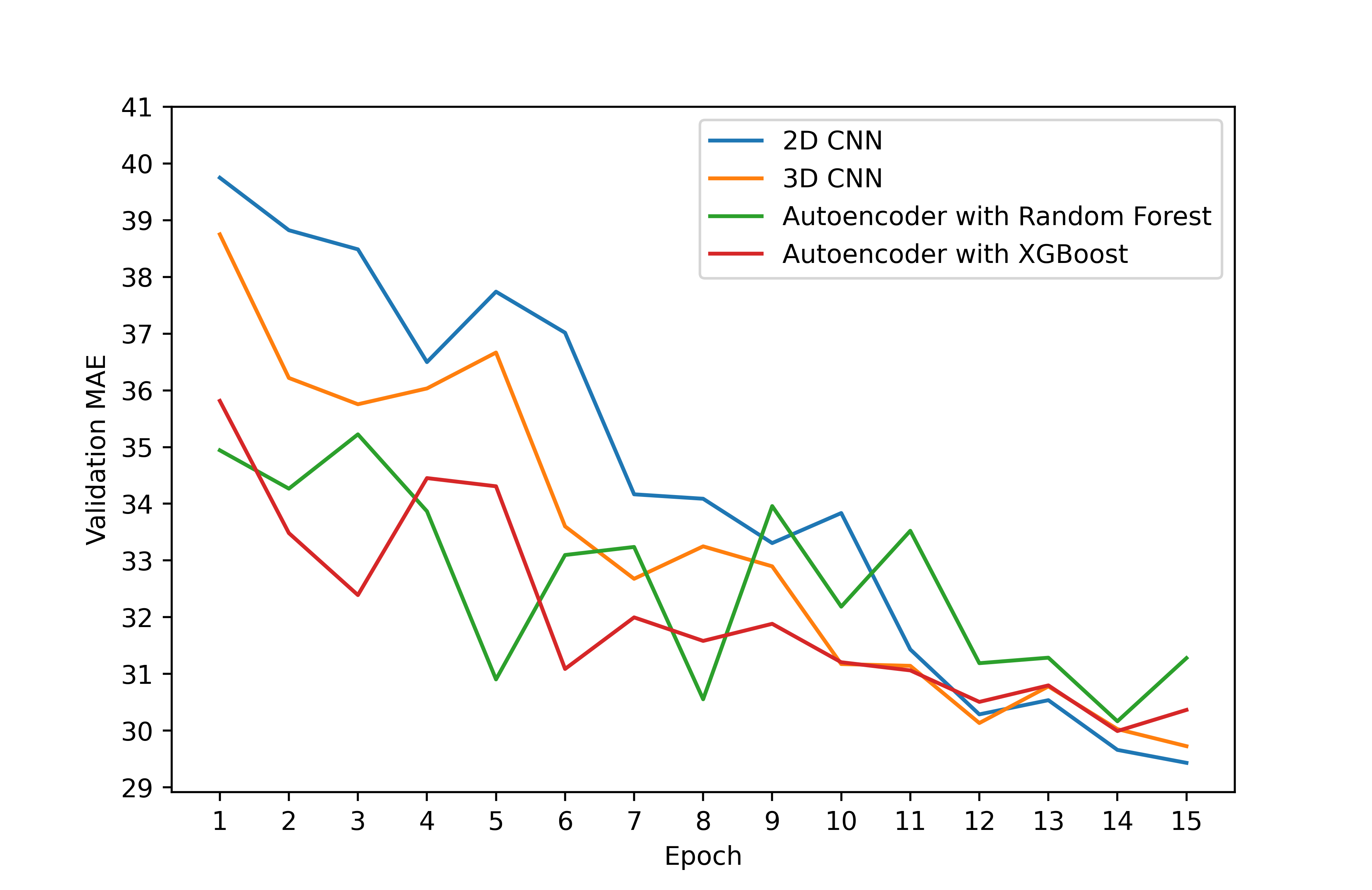}
\caption{Model-wise plots of validation MAE for one of the 12 folds.}
\label{maeplots}
\end{figure}
\\
Another crucial observation we made during training was that for all architectures (and in almost all 12 runs of cross-validation), the training loss vs. epochs graphs were highly volatile, which seems to be a consequence of the batch size we chose. This is corroborated by Fig.~\ref{maeplots}, which displays the epoch-wise perfeormance our models achieved on one of the 12 folds used for validation. The fact that Pytorch does not allow images of different dimensions in a single batch forced us to use a batch size of 1. Essentially, this annulled the concept of batch training in our work. Due to this, we could not add batch normalization, which resulted in slower convergence of our models. It was also observed that despite adding multiple layers of dropouts, our models were somewhat prone to overfitting, possibly because we omitted batch normalization layers in our architecture and because of the small size of the dataset, containing only about 3600 data points. 
\\
\section{Conclusion}

In this paper, we proposed a way to predict the auditory attention score from EEG signals of a person using computer vision based architectures. In order to apply deep learning algorithms, we converted the time-series data into images using a technique known as GADF. We used adaptive maxpooling in our architectures to combat the varying sizes of images. It was found that 2D CNN gave the best results among all the models, in fact even bettering existing performances on the chosen dataset~\cite{bajaj2020}.
 
We plan to extend our work by training the models on a larger dataset, which will give us more scope to experiment and improve the generalizability of our models. We plan to train much deeper 3D CNN models (with better memory resources at hand) and advanced architectures such as 2D CNNs + LSTMs that would help us better capture interchannel information. We also plan to use better data (signal) preprocessing techniques before converting the data into images, allowing models to extract features better and consequently, learn faster. We hope this work of ours serves as a platform to improvise and innovate further and thereby contributing to the research community.

\bibliographystyle{ieeetr}
% \balance
\bibliography{bibliography.bib}
\end{document}